\documentclass[conference]{IEEEtran}

\usepackage{cite}
\usepackage{amsmath,amssymb,amsfonts}
\usepackage{graphicx}
\usepackage{textcomp}
\usepackage{xcolor}
\usepackage{booktabs}
\usepackage{multirow}
\usepackage{subcaption}
\usepackage{url}
\usepackage{fancyhdr}

\def\BibTeX{{\rm B\kern-.05em{\sc i\kern-.025em b}\kern-.08em
    T\kern-.1667em\lower.7ex\hbox{E}\kern-.125emX}}

\begin{document}

\title{DEM Refinement and Validation on the Lunar Surface Using Shape-from-Shading with Chandrayaan-2 OHRC Imagery}

\author{
\begin{tabular}{ccc}
    \begin{minipage}[t]{0.3\textwidth}
        \centering
        1\textsuperscript{st} Aaranay Aadi \\
        \textit{School of Computer Science and Engineering,} \\
        \textit{Manipal University Jaipur} \\
        Jaipur, India \\
        aaranayaadi@gmail.com
    \end{minipage} & 
    \begin{minipage}[t]{0.3\textwidth}
        \centering
        2\textsuperscript{nd} Jai Gopal Singla \\
        \textit{Space Applications Centre (SAC),} \\
        \textit{Indian Space Research Organisation (ISRO)} \\
        Ahmedabad, India \\
        jaisingla@sac.isro.gov.in
    \end{minipage} & 
    \begin{minipage}[t]{0.3\textwidth}
        \centering
        3\textsuperscript{rd} Nitant Dube \\
        \textit{Space Applications Centre (SAC),} \\
        \textit{Indian Space Research Organisation (ISRO)} \\
        Ahmedabad, India \\
        nitant@sac.isro.gov.in
    \end{minipage}
\end{tabular}
}

\fancypagestyle{firstpage}{
  \fancyhf{} 
  \renewcommand{\headrulewidth}{0pt} 
  \fancyfoot[C]{\footnotesize This work is a preprint and has not been peer-reviewed. It has been posted to share preliminary findings. The views expressed here are those of the authors and do not necessarily reflect the views or policies of ISRO SAC.}
}

\pagestyle{empty} 

\maketitle
\thispagestyle{firstpage} 
\pagestyle{empty}         

\begin{abstract}
This study presents a Shape from Shading (SfS) framework to enhance sub-metre resolution lunar digital elevation models (DEMs) using imagery from the Orbiter High Resolution Camera (OHRC) aboard Chandrayaan-2. The framework applies SfS to an independent OHRC image of the same region, enabling SfS not just as a refinement tool, but as a source of new topographic data, unconstrained by stereo baseline limitations. The method is applied across three lunar sites, including the Cyrillus crater, the Vikram landing region, and the lunar south pole (Mons Mouton), with a systematic three-stage parameter sweep on the SfS smoothness weight. Results show measurable topographic enhancement, particularly in surface slope statistics, revealing fine-scale crater morphology previously unresolved. A limiting case is also characterized, where large pitch angle separation between the shading image and stereo pair reduces SfS sensitivity, and partial footprint coverage of the shading image is identified as a factor influencing spatially variable enhancement quality.
\end{abstract}


\section{Introduction}

Accurate topographic models of the lunar surface are a necessity for geological
mapping, landing-site hazard assessment, and terrain study. Orbital stereophotogrammetry remains the primary
means of constructing such models, but the recoverable spatial frequency
content is limited by sensor ground sampling distance, stereo baseline
geometry, and image matching performance in featureless or
shadow-dominated terrain. Shape from Shading (SfS) is a method that estimates surface shape from image brightness under known illumination conditions. When applied to an additional image with a different viewing and lighting geometry, it provides extra surface detail that is not captured by stereo reconstruction, enabling refinement at finer spatial scales than the stereo baseline allows~\cite{horn1970}.

India's Chandrayaan-2 mission carries the Orbiter High Resolution Camera
(OHRC), which images the lunar surface at 0.25\,m/pixel from a
100\,km orbit, making it one of the highest-resolution lunar orbital
datasets. Standard photogrammetric reconstruction methods are made possible by the ISIS~\cite{isis2023} and Ames Stereo Pipeline (ASP)~\cite{beyer2018} toolchain, which supports the processing of lunar data. This work builds
upon our earlier extension of these tools to OHRC data~\cite{arxiv_ohrc}
and further develop the pipeline to incorporate Shape-from-Shading (SfS)
for surface refinement.

\section{Related Work and Background}

\subsection{Related Work}

Shape-from-Shading (SfS), first formalized by Horn~\cite{horn1970} under a Lambertian reflectance model, aims to recover surface shape from image irradiance. However, this formulation is ill-posed and sensitive to noise, requiring additional constraints for stability.

Lohse \textit{et al.}~\cite{lohse2006} applied SfS to planetary surfaces, addressing non-Lambertian effects~\cite{hapke2012} and low illumination angles, which are critical for lunar and planetary regolith. The Lunar-Lambertian model~\cite{mcewen1991} became a standard to capture these effects, particularly for airless bodies like the Moon.

Recent advancements have integrated SfS with stereo-derived geometry to reduce ambiguity. Alexandrov and Beyer~\cite{alexandrov2018} introduced a multiview SfS framework within the Ames Stereo Pipeline (ASP), successfully applied to LROC NAC imagery~\cite{robinson2010}, especially in polar regions with challenging illumination conditions.

Hybrid stereo-photoclinometry approaches further enhance reconstruction fidelity by using SfS as a refinement step. These methods emphasize the critical role of acquisition geometry, that are illumination and viewing angles, in determining the quality of SfS reconstructions.

\subsection{Shape from Shading: Theory}
The SfS problem estimates the surface height \(z(\mathbf{x})\) from observed image irradiance \(I(\mathbf{x})\), given a known reflectance model \(R\) and illumination direction \(\hat{\ell}\). It relates image brightness to surface orientation through the image irradiance equation~\cite{ikeuchi1981numerical}:
\begin{equation}
  I(\mathbf{x}) = R\!\left(\hat{n}(\mathbf{x}),\,\hat{\ell},\,\hat{v}\right),
  \label{eq:irradiance}
\end{equation}
where \(\hat{n}\) is the surface normal and \(\hat{v}\) is the viewing direction.

This work uses the Lunar-Lambertian reflectance model~\cite{mcewen1991}, which is widely used for airless planetary surfaces:
\begin{equation}
  R_{\mathrm{LL}} =
  A\!\left[\frac{2\cos\theta_i}{\cos\theta_i+\cos\theta_e}
  + (1-A)\cos\theta_i\right],
  \label{eq:lunar_lambert}
\end{equation}
where the incidence and emission angles are defined as
\begin{equation}
  \cos\theta_i = \hat{n} \cdot \hat{\ell},
  \qquad
  \cos\theta_e = \hat{n} \cdot \hat{v}.
  \label{eq:angles}
\end{equation}

The surface normal is expressed in terms of the height gradients as
\begin{equation}
  \hat{n} = \frac{(-p,\,-q,\,1)^{\!\top}}{\sqrt{1+p^2+q^2}},
  \qquad
  p = \frac{\partial z}{\partial x},\;\;
  q = \frac{\partial z}{\partial y}.
  \label{eq:normal_grad}
\end{equation}

The SfS problem is ill-posed and therefore requires regularization. In the ASP formulation~\cite{alexandrov2018}, this is addressed by combining a data-fidelity term with a smoothness constraint and an optional constraint to remain close to an initial DEM:
\begin{align}
  \mathcal{E}(z) &=
    \int_\Omega \left[I(\mathbf{x}) - R(\hat{n}(z))\right]^2 d\mathbf{x}
    \label{eq:sfs_obj} \\
  &+ W \int_\Omega \left\|\nabla^2 z(\mathbf{x})\right\|^2 d\mathbf{x}
    \nonumber \\
  &+ C \int_\Omega \left[z(\mathbf{x}) - z_0(\mathbf{x})\right]^2 d\mathbf{x},
    \nonumber
\end{align}
where \(W\) controls surface smoothness and \(C\) controls the strength of the constraint to the initial DEM. As \(W \to \infty\), the solution becomes increasingly smooth and approaches \(z_0\). As \(W \to 0\), the solution relies more strongly on image shading, enabling finer detail recovery but increasing sensitivity to noise.

\subsection{SfS at the Lunar South Pole}

At the lunar south pole, the Sun remains perpetually within
$\sim$1.5$^\circ$ of the horizon, producing solar incidence angles
$\theta_i > 82^\circ$. Under such grazing illumination, small
topographic undulations generate large brightness contrasts, rendering
$\partial I / \partial z$ large and the SfS problem
well-conditioned~\cite{alexandrov2018,zuber2012}. Conversely, the same
geometry causes deep, pervasive shadows that undermine stereo matching
by eliminating texture in occluded regions. SfS is therefore
particularly well-suited as a \emph{complement} to stereo in polar
terrain.

\section{Contribution}

This work extends the ASP SfS framework to high-resolution OHRC data. Key contributions include: (i) an end-to-end SfS pipeline for Chandrayaan-2 OHRC imagery, generating SfS enhanced DEMs at sub-metre resolution; (ii) an SfS formulation where the shading image is independent of the stereo pair, allowing additional observations; (iii) a systematic exploration of the smoothness weight parameter to identify stable regimes for OHRC-based SfS; (iv) application across multiple lunar sites, including a south pole region; (v) quantitative validation against OHRC stereo DEMs and NAC DTMs, assessing elevation consistency and surface refinement; and (vi) analysis of the influence of acquisition geometry on SfS performance, particularly angular separation and footprint overlap, to analyze limitations.

\section{Datasets}

Table~\ref{tab:datasets} lists the three OHRC image triplets used in
this study. For each site, $\mathcal{I}_1$ and $\mathcal{I}_2$ form the
stereo pair from which $z_0$ is derived; $\mathcal{I}_3$ is the
independent image used as the SfS shading source. Key photometric
metadata, that are solar incidence angle $\theta_i$, solar azimuth
$\phi_\odot$, and pitch angle $\theta_p$, are listed for
each image, as these quantities directly govern stereo quality and SfS
conditioning.

\begin{table*}[t]
  \caption{OHRC Image Datasets Used in This Study~\cite{isro_ch2_pradan}. All images are
           panchromatic, 0.20-0.26\,m/pixel. Image IDs, dates, and exact
           angular values are mentioned. $\Delta\theta_i$ and $\Delta\theta_p$ are
           computed with respect to the mean stereo geometry and the third images. All Image IDs share the common prefix: \texttt{ch2\_ohr\_nrp\_}}
  \label{tab:datasets}
  \centering
  \begin{tabular}{p{3.2cm} l l l c c c c c c}
    \toprule
    \textbf{Site} &
    \textbf{Dataset} &
    \textbf{Image ID} &
    \textbf{Role} &
    \textbf{Acq.\ Date} &
    $\theta_i$ ($^\circ$) &
    $\phi_\odot$ ($^\circ$) &
    $\theta_p$ ($^\circ$) &
    $\Delta\theta_i$ &
    $\Delta\theta_p$ \\
    \midrule

    \multirow{3}{3cm}{\raggedright Vikram landing\\region}
      & \multirow{3}{*}{A}
      & 20240425T1209509264 & Stereo $\mathcal{I}_1$ & 25-04-2024 & 79.155330 & 304.085609 & -10.728805 & --- & --- \\
      & & 20240425T1406019344 & Stereo $\mathcal{I}_2$ & 25-04-2024 & 78.470478 & 303.869837 & 9.639808 & --- & --- \\
      & & 20240425T1603031918 & SfS $\mathcal{I}_3$ & 25-04-2024 & 79.097263 & 304.447986 & -12.289252 & 0.28 & 11.75 \\

    \midrule

    \multirow{3}{3cm}{\raggedright South pole\\(Mons Mouton)}
      & \multirow{3}{*}{B}
      & 20241117T1637538832 & Stereo $\mathcal{I}_1$ & 17-11-2024 & 84.707470 & 307.104947 & -17.127440 & --- & --- \\
      & & 20241117T1439153327 & Stereo $\mathcal{I}_2$ & 17-11-2024 & 86.216884 & 301.693569 & 17.278636 & --- & --- \\
      & & 20241215T1104345070 & SfS $\mathcal{I}_3$ & 15-12-2024 & 82.374290 & 334.899255 & -18.447503 & 3.09 & 18.53 \\

    \midrule

    \multirow{3}{3cm}{Cyrillus crater\\(limiting case)}
      & \multirow{3}{*}{C}
      & 20210401T2200364910 & Stereo $\mathcal{I}_1$ & 01-04-2021 & 80.206470 & 271.202165 & -18.500424 & --- & --- \\
      & & 20210402T0155096873 & Stereo $\mathcal{I}_2$ & 02-04-2021 & 80.003112 & 270.902984 & 10.685380 & --- & --- \\
      & & 20231004T0406038822 & SfS $\mathcal{I}_3$ & 04-10-2023 & 80.511339 & 272.874656 & 26.599231 & 0.41 & 30.51 \\

    \bottomrule
  \end{tabular}
\end{table*}

\section{DEM initialization from OHRC Imagery}

\subsection{Image Selection and Three-Image Strategy}

Three OHRC images of the target region are selected. Images
$\mathcal{I}_1$ and $\mathcal{I}_2$ form the stereo pair used to
construct the initial DEM $z_0$, while $\mathcal{I}_3$ is an
independent acquisition used exclusively for SfS shading.

To ensure compatibility with the SfS domain, $\mathcal{I}_3$ is
mapprojected~\cite{lapaine2016map} onto $z_0$:
\begin{equation}
  \mathcal{I}_3^{\mathrm{mp}}(\mathbf{x}) =
  \mathcal{I}_3\!\left(\pi\!\left(z_0(\mathbf{x}),\,\mathbf{P}_3\right)\right),
  \label{eq:mapproject}
\end{equation}
where $\pi(\cdot,\mathbf{P}_3)$ denotes projection through the camera
model of $\mathcal{I}_3$. Spatial overlap is verified via the valid
footprint of $\mathcal{I}_3^{\mathrm{mp}}$.

\subsection{Stereo Reconstruction}

Stereo reconstruction follows our prior OHRC photogrammetry pipeline~\cite{arxiv_ohrc}, including SPICE-based sensor modeling, bundle adjustment, and dense matching.

\subsection{DEM Registration and Preparation}

The reconstructed DEM is vertically aligned to a reference NAC DTM by
estimating a constant offset, corresponding to a simplified
least-squares surface registration with translation-only degrees of
freedom~\cite{besl1992}:
\begin{equation}
  \Delta z^* = \arg\min_{\Delta z}\,
  \frac{1}{|\mathcal{S}|}\sum_{\mathbf{x}\in\mathcal{S}}
  \left[z_0(\mathbf{x}) + \Delta z - z_{\mathrm{ref}}(\mathbf{x})\right]^2,
  \label{eq:pc_align}
\end{equation}
where $\mathcal{S}$ denotes the set of valid overlapping pixels. This
assumes prior horizontal co-registration between the two DEMs, such
that residual misalignment is dominated by a constant vertical bias.

\begin{figure}[htbp]
  \centering
  \includegraphics[width=\columnwidth]{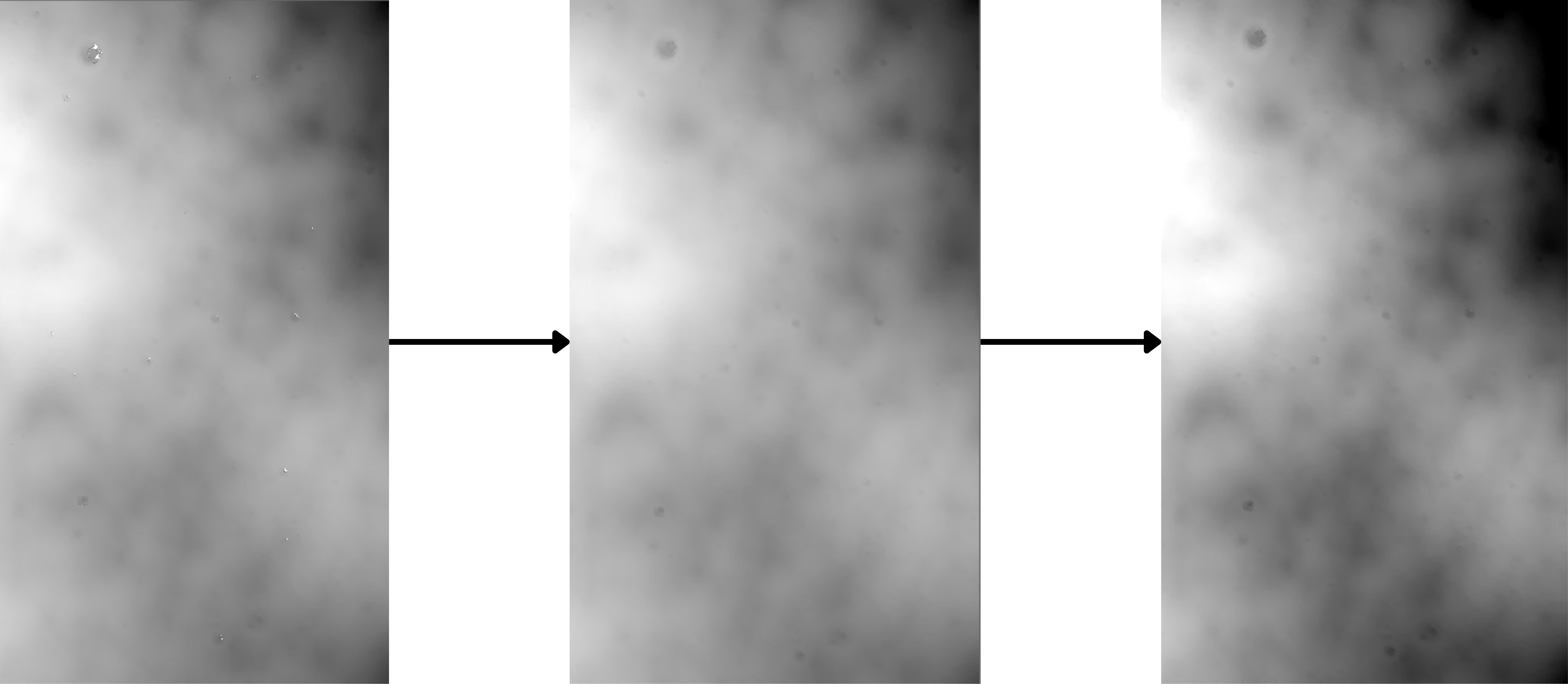}
  \caption{Iterative SfS production stages. (a) Cropped stereo DEM with
    no-data gaps. (b) Gap-filled DEM after blending with NAC DTM.
    (c) SfS-enhanced DEM $z_{\mathrm{SfS}}$, showing recovery of
    fine-scale topographic detail from $\mathcal{I}_3$.}
  \label{fig:iterative_sfs}
\end{figure}

No-data gaps are filled using priority blending, preserving OHRC-derived
values wherever available and using the NAC DTM elsewhere. For the regions where NAC acquisition is unavailable, interpolation~\cite{perez2016comparison} is used to fill the gaps. The resulting
DEM serves as the input to SfS refinement. The processing workflow, including DEM preparation, gap filling, and SfS refinement, is illustrated in Fig.~\ref{fig:iterative_sfs}.

\section{Shape from Shading: Methodology and Parameter Study}

\begin{figure*}[t] 
  \centering
  \includegraphics[width=\textwidth]{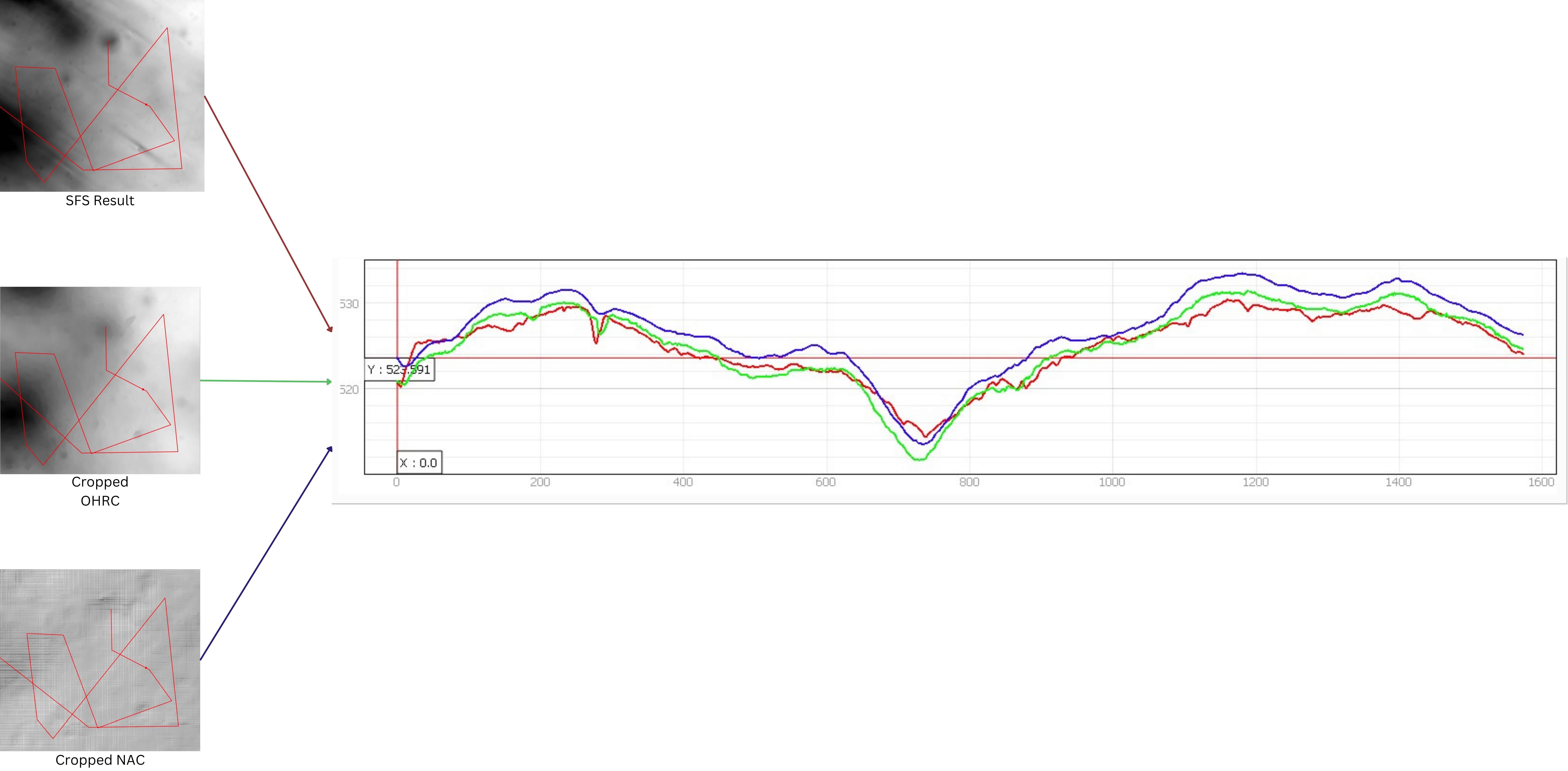}
  \caption{Terrain profile comparison over a transect in the Vikram
    landing region. Horizontal axis: distance along profile
transect (m). Vertical axis: elevation (m). The SfS-enhanced profile (red) resolves
    sub-resolution topographic undulations relative to both the stereo
    DEM $z_0$ (green) and the NAC DTM reference (blue), while
    remaining consistent with the large-scale trend of the reference.
    Deviations between $z_0$ and the NAC DTM reflect the higher spatial
    resolution of the OHRC stereo product.}
  \label{fig:profile_vikram}
\end{figure*}

\subsection{SfS on the Independent Third Image}

SfS is applied to image $\mathcal{I}_3$ using $z_0$ as the initial
surface. This design confers two advantages. First, the SfS shading
signal is entirely independent of the stereo reconstruction, so any
topographic detail recovered by SfS is genuinely new information not
already encoded in $z_0$. Second, the solar geometry of $\mathcal{I}_3$
can be chosen independently of the stereo pair; for polar sites, a pass
with near-grazing illumination is preferred for SfS (maximizing
topographic sensitivity via large $\partial R / \partial \theta_i$),
while a pass with higher solar elevation is preferred for stereo
(improving texture and minimizing shadows).

The variational objective~(\ref{eq:sfs_obj}) is minimized iteratively.
At each iteration, the rendered brightness $R(\hat{n}(z^{(k)}))$ is
compared to $\mathcal{I}_3$, and the height field is updated by gradient
descent with step size adapted to the local curvature of $\mathcal{E}$.
Camera corrections (small perturbations to $\mathbf{R}$ and
$\mathbf{t}$) are jointly optimized to absorb residual pointing errors
that would otherwise masquerade as topographic signal.

\subsection{Parameter Sweep: Smoothness Weight $W$}

The smoothness weight $W$ in~(\ref{eq:sfs_obj}) is the primary
free parameter controlling the tradeoff between shading fidelity and
surface regularity. Three sweep stages are conducted.

\textbf{Stage 1 --- Decade sweep.}
$W$ is sampled at $\{10^4,\,10^3,\,10^2,\,10^1\}$ with baseline
secondary parameters ($C = 10^{-3}$, $N = 10$) where $C$ is the  initial-DEM constraint
weight and $N$ is the number of maximum iterations. This identifies the
qualitative operating regime: the minimum $W$ at which the SfS output
remains free of noise artefacts and the maximum $W$ at which
shading-driven modulation is still visible.

\textbf{Stage 2 --- Intra-decade refinement.}
For each decade $D$ exhibiting visible topographic modulation, $W$ is
further sampled at $\{D,\;0.75D,\;0.5D,\;0.25D\}$ to resolve the
transition point between productive enhancement and over-smoothing
or noise amplification.

\textbf{Stage 3 --- Interaction study.}
The optimal $W^*$ from Stage 2 is held fixed while $C \in
\{10^{-1},\,10^{-2},\,10^{-3},\,10^{-4}\}$ and $N \in \{5,10,20,50\}$
are varied independently to characterize their secondary influence.

\subsection{Evaluation Metrics}

Let $z_{\mathrm{SfS}}$ denote the converged SfS surface and $z_0$ the
initial DEM. Three metrics are reported.

\textbf{(i) Slope standard deviation.}
\begin{equation}
  \sigma_s = \mathrm{std}\!\left(
    \arctan\!\sqrt{\left(\frac{\partial z}{\partial x}\right)^2
                 +\left(\frac{\partial z}{\partial y}\right)^2}
  \right),
  \label{eq:sigma_s}
\end{equation}
computed from both $z_0$ and $z_{\mathrm{SfS}}$. An increase
$\Delta\sigma_s > 0$ indicates recovery of additional topographic
relief.

\textbf{(ii) Height difference map.}
$\delta z(\mathbf{x}) = z_{\mathrm{SfS}}(\mathbf{x}) - z_0(\mathbf{x})$
localises the spatial distribution of SfS-induced modifications.
Modifications concentrated at geomorphologically coherent features
(crater rims, ridge crests) validate that SfS is recovering genuine
topography rather than noise.

\textbf{(iii) Photometric residual.}
\begin{equation}
  \varepsilon_R = \frac{1}{|\Omega|}
    \int_\Omega \left[I(\mathbf{x}) - R\!\left(\hat{n}(z_{\mathrm{SfS}})\right)
    \right]^2 d\mathbf{x},
  \label{eq:photo_residual}
\end{equation}
which quantifies how well the final surface explains the observed
image brightness; a lower value indicates better shading reproduction.

\section{Results}

\subsection{Vikram Landing Region: Profile-Based Validation}
\begin{figure}[htbp]
  \centering
  \begin{subfigure}[b]{0.48\columnwidth}
     \includegraphics[width=\linewidth]{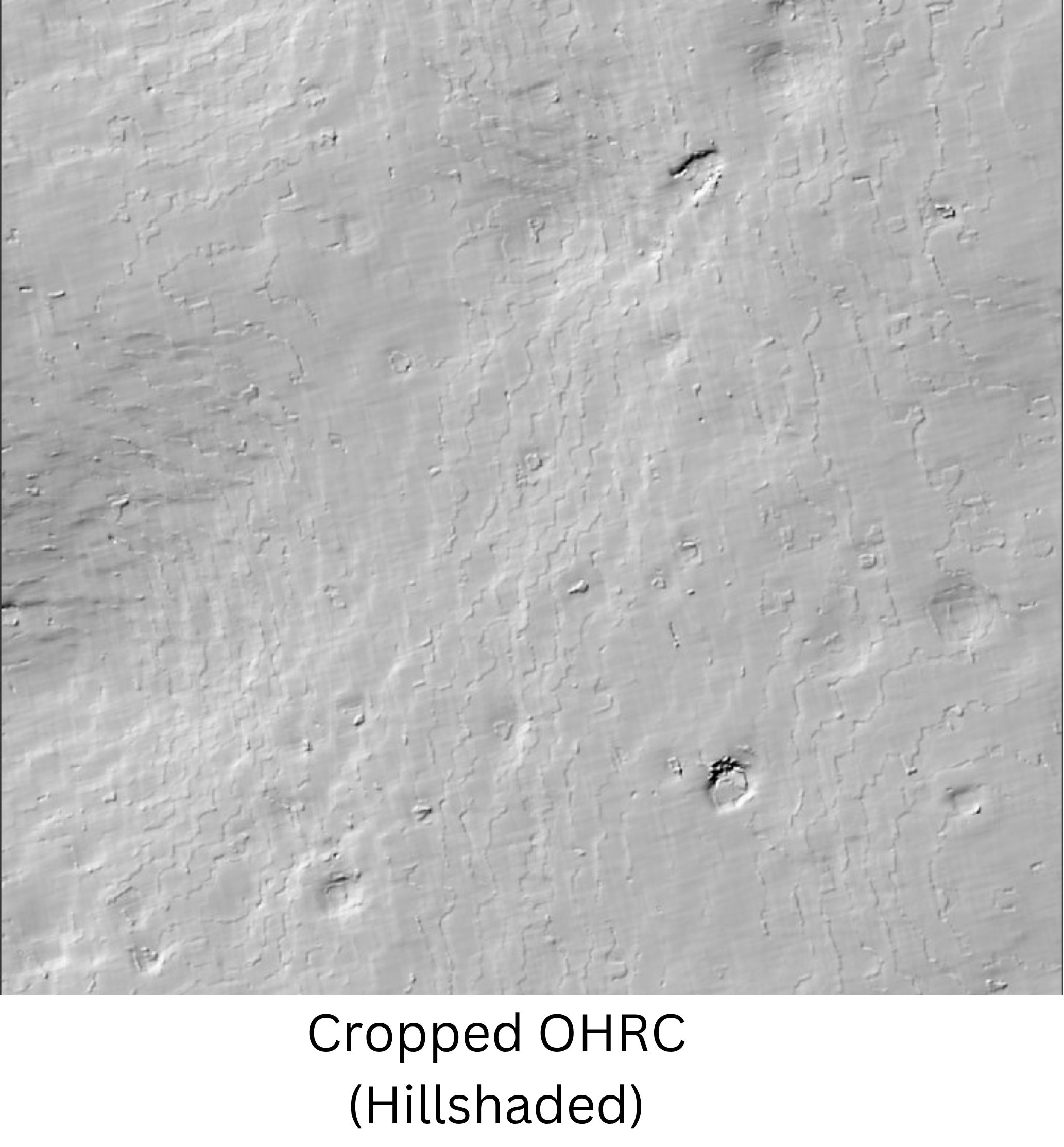}
    \caption{Initial DEM ($z_0$)}
    \label{fig:hs_vikram_initial}
  \end{subfigure}
  \hfill
  \begin{subfigure}[b]{0.48\columnwidth}
    \includegraphics[width=\linewidth]{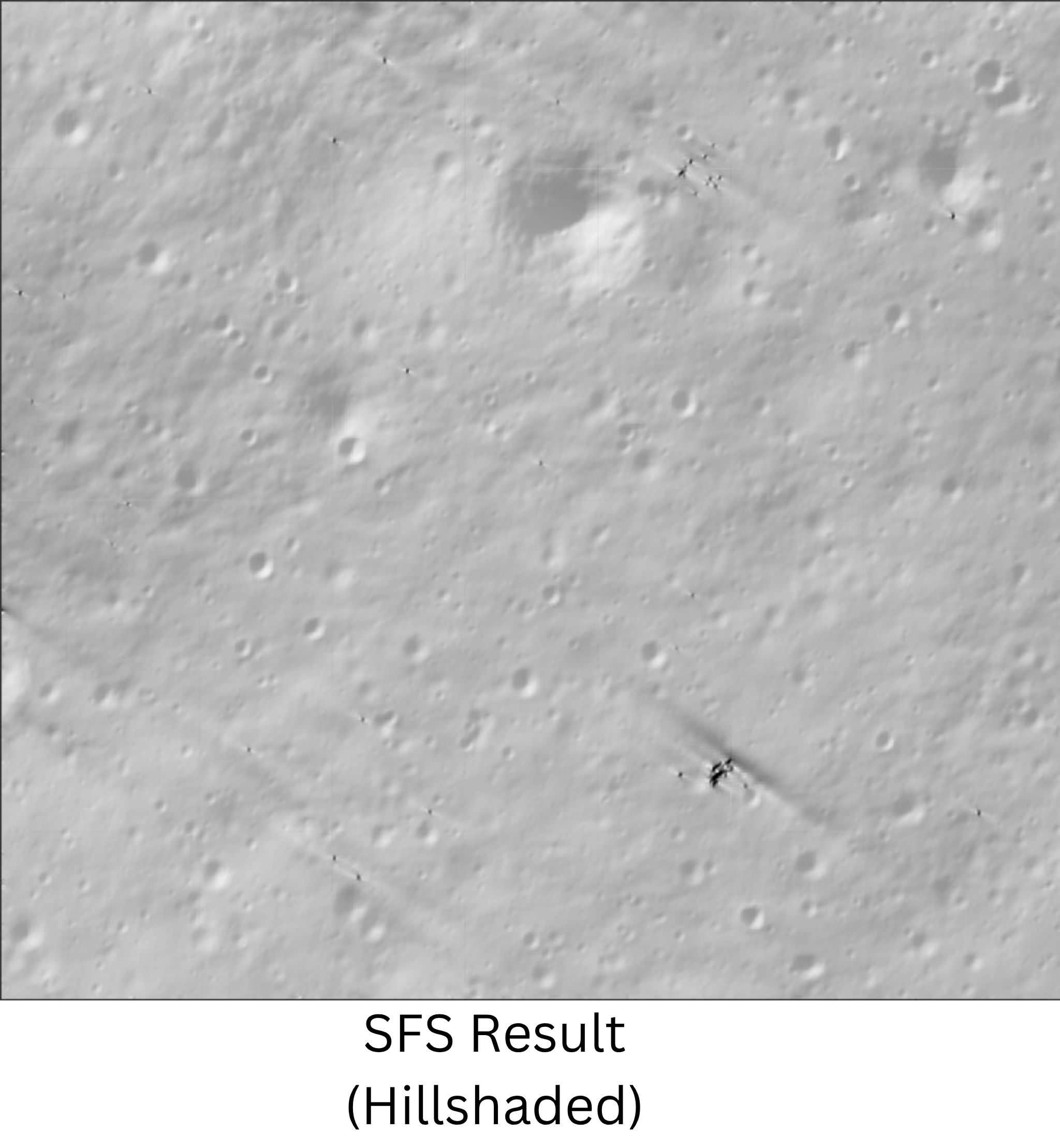}
    \caption{SfS DEM ($z_{\mathrm{SfS}}$)}
    \label{fig:hs_vikram_sfs}
  \end{subfigure}
  \caption{Hillshaded DEMs of the Vikram landing region before (left)
    and after (right) SfS enhancement ($W = 50$, $C = 10^{-3}$,
    $N = 8$). Illumination azimuth fixed at $315^\circ$ for both
    panels. Fine-scale crater morphology and metre-scale slope
    variations absent in $z_0$ are resolved in $z_{\mathrm{SfS}}$.}
  \label{fig:hillshade_vikram}
\end{figure}

As shown in Fig.~\ref{fig:profile_vikram}, the profile comparisons of the stereo OHRC DEM, SfS result, and NAC DTM are represented by the green, red, and blue lines, respectively. These profiles highlight subtle depth differences, particularly in less illuminated regions. Fig.~\ref{fig:hillshade_vikram} displays the variations between the two DEMs under hillshading conditions. 

\subsection{SfS Parameter Sweep}

Table~\ref{tab:params} presents the sensitivity analysis of SfS parameters.

\begin{table}[htbp]
  \caption{SfS Parameter Sensitivity: Slope Std.\ Dev.\ Change
    Relative to Initial DEM (Dataset A, Vikram region)}
  \label{tab:params}
  \centering
  \begin{tabular}{cccc}
    \toprule
    $W$ & $C$ & $N$ & $\Delta\sigma_s$ \\
    \midrule
    $10^4$ & $10^{-3}$ & 8  & Baseline          \\
    $10^3$ & $10^{-3}$ & 8  & $+4.2\%$          \\
    $10^2$ & $10^{-3}$ & 8  & $+18.1\%$         \\
    80     & $10^{-3}$ & 8  & $+21.3\%$         \\
    50     & $10^{-3}$ & 8  & $\mathbf{+22.9\%}$ \\
    25     & $10^{-3}$ & 8  & $+23.4\%$ (noisy) \\
    $10^1$ & $10^{-3}$ & 8  & $+23.7\%$ (noisy) \\
    80     & $10^{-2}$ & 8  & $+19.8\%$         \\
    80     & $10^{-4}$ & 8  & $+22.1\%$         \\
    \bottomrule
  \end{tabular}
\end{table}

\textbf{Decade Sweep (Stage 1):}  
At $W = 10^4$, the SfS output closely matched the initial DEM, dominated by smoothness regularization. As $W$ decreased to $10^2$, more topographic detail emerged, with $\sigma_s$ increasing by $18.1\%$. However, further reduction to $W=10$ introduced noise, indicating over-sensitivity to radiometric errors.

\textbf{Intra-decade Refinement (Stage 2):}  
Within the $W = 10^2$ range, the optimal value was $W = 50$, balancing structural enhancement with noise suppression ($\Delta\sigma_s = 22.9\%$). Lower values ($W < 25$) introduced noise, while higher values ($W > 80$) resulted in under-enhancement, suggesting $W \approx 50$ as the most stable choice for OHRC SfS.

\textbf{Secondary Parameter Sensitivity (Stage 3):}  
Increasing the regularization weight $C$ above $10^{-2}$ suppressed SfS detail, while values below $10^{-4}$ led to drift. A stable value of $C = 10^{-3}$ was chosen for absolute height consistency while preserving fine-scale detail. Iteration tests showed that solutions converged rapidly within 8–10 iterations, with no significant improvement at $N > 10$.

\textbf{Saturation and Noise:}  
Although $\Delta\sigma_s$ increased slightly for $W < 50$, this was due to noise amplification rather than meaningful topographic improvement, indicating that the SfS solution saturated beyond this point.

\subsection{South Pole: Mons Mouton Terrain Profiles}

\begin{figure}[h]
  \centering
  \includegraphics[width=\linewidth]{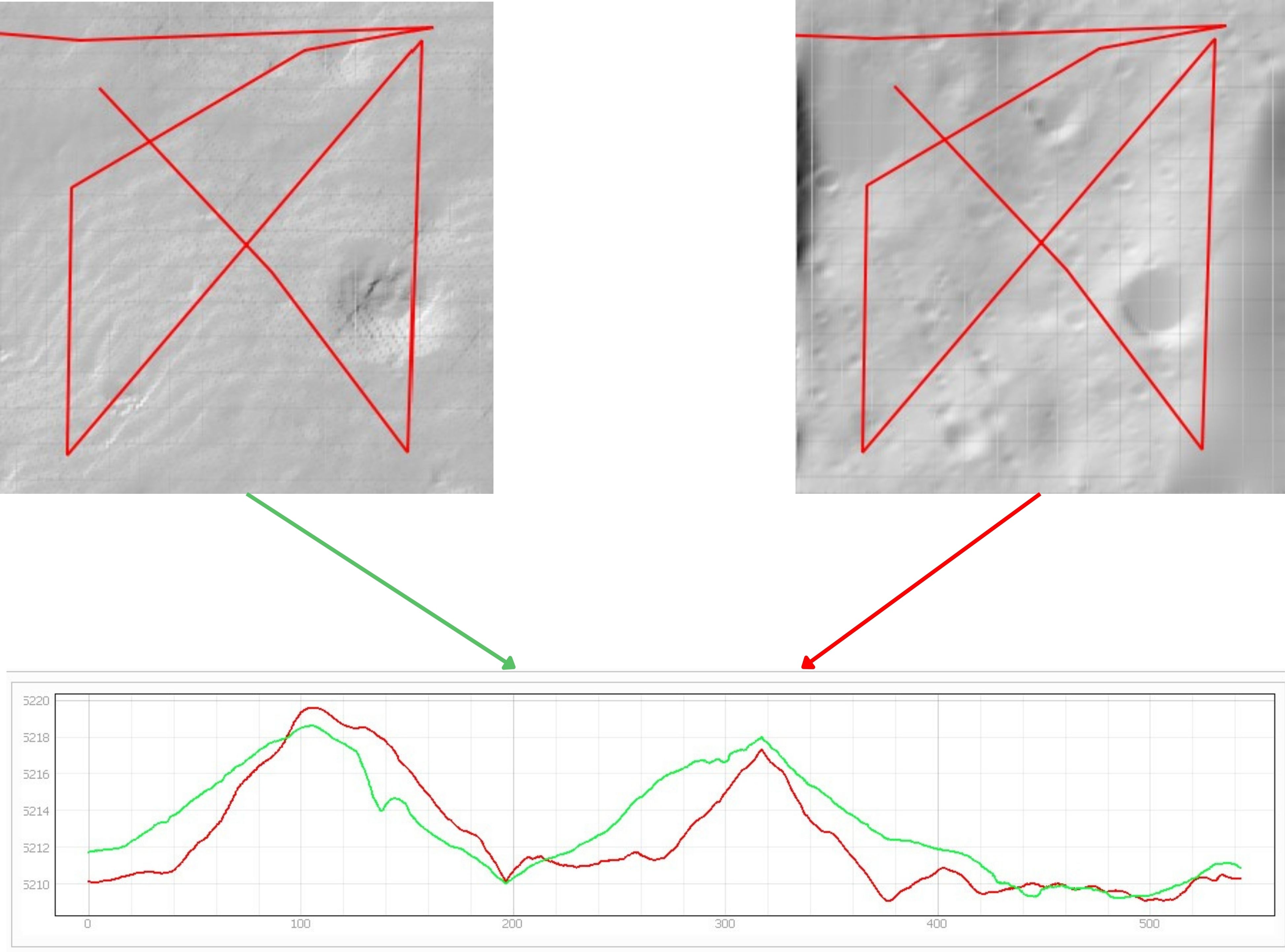}
  \caption{Terrain profile comparison at Mons Mouton ($84.289^\circ$S, $32.808^\circ$E) for a small-region transect. The plot compares $z_0$ (blue) vs $z_{\mathrm{SfS}}$ (red), showing consistent SfS enhancement and topographic modulation across the region.}
  \label{fig:profile_sp_small}
\end{figure}

\begin{figure}[h]
  \centering
  \includegraphics[width=\linewidth]{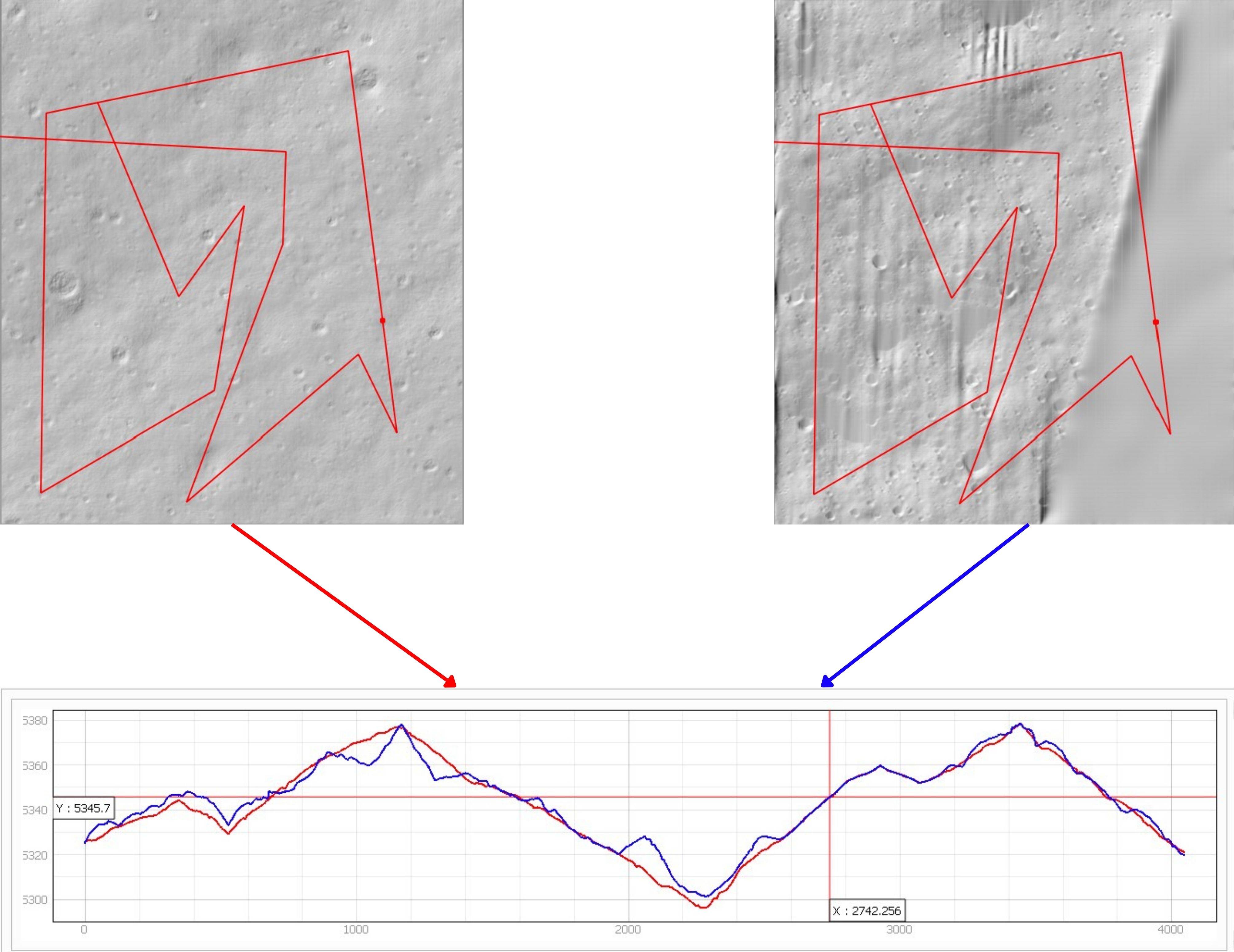}
  \caption{Terrain profile comparison at Mons Mouton for a larger-region transect. The red line represents the initial DEM, while the blue line represents the SFS DEM. The right portion shows reduced SfS modulation corresponding to the region outside the $\mathcal{I}_3$ footprint coverage.}
  \label{fig:profile_sp_large}
\end{figure}

The south polar target is centred on Mons Mouton
($84.289^\circ$S, $32.808^\circ$E), a flat-topped mountain whose
elevated summit provides near-continuous illumination and makes it
a site of significant scientific and exploratory interest. The steep
flanks and summit plain of Mons Mouton present a demanding
topographic test for SfS: the flat-topped geometry should yield a
suppressed SfS response on the plateau while producing strong
enhancement on the surrounding slopes, providing a natural
morphological control. Fig.~\ref{fig:profile_sp_small} and Fig.~\ref{fig:profile_sp_large} demonstrate the results at Mons Mouton.

\section{Discussions}
\label{sec:discussion}

\textbf{SfS as a new data source.}
The use of a third independent image $\mathcal{I}_3$ sets this work apart from traditional post-processing enhancements. Acquired at a different solar geometry, $\mathcal{I}_3$ captures topographic features absent from the stereo point cloud. As a result, the SfS solution $z_{\mathrm{SfS}}$ is not merely a smoothed $z_0$, but a surface that explains an independent physical observation. This is demonstrated in the profile comparisons (Figs.~\ref{fig:profile_vikram}, \ref{fig:profile_sp_small}, and \ref{fig:profile_sp_large}), where SfS recovers topographic features that are neither present in the NAC DTM nor the stereo DEM.

\textbf{Footprint coverage as a spatial quality predictor.}
Figure~\ref{fig:profile_sp_large} shows a zone of reduced SfS modulation in the Mons Mouton transect, corresponding to areas where $\mathcal{I}_3$ provides no valid coverage after maprojection onto $z_0$. In these regions, the SfS objective reduces to a balance between the smoothness term and the initial DEM, causing the solution to relax toward $z_0$. This highlights the need to verify $\mathcal{I}_3$ footprint coverage \emph{before} defining the SfS crop boundary, as discussed in Section~V-A.

\begin{figure}[htbp]
  \centering
  \begin{subfigure}[b]{0.48\columnwidth}
    \includegraphics[width=\linewidth, height=4cm, keepaspectratio]{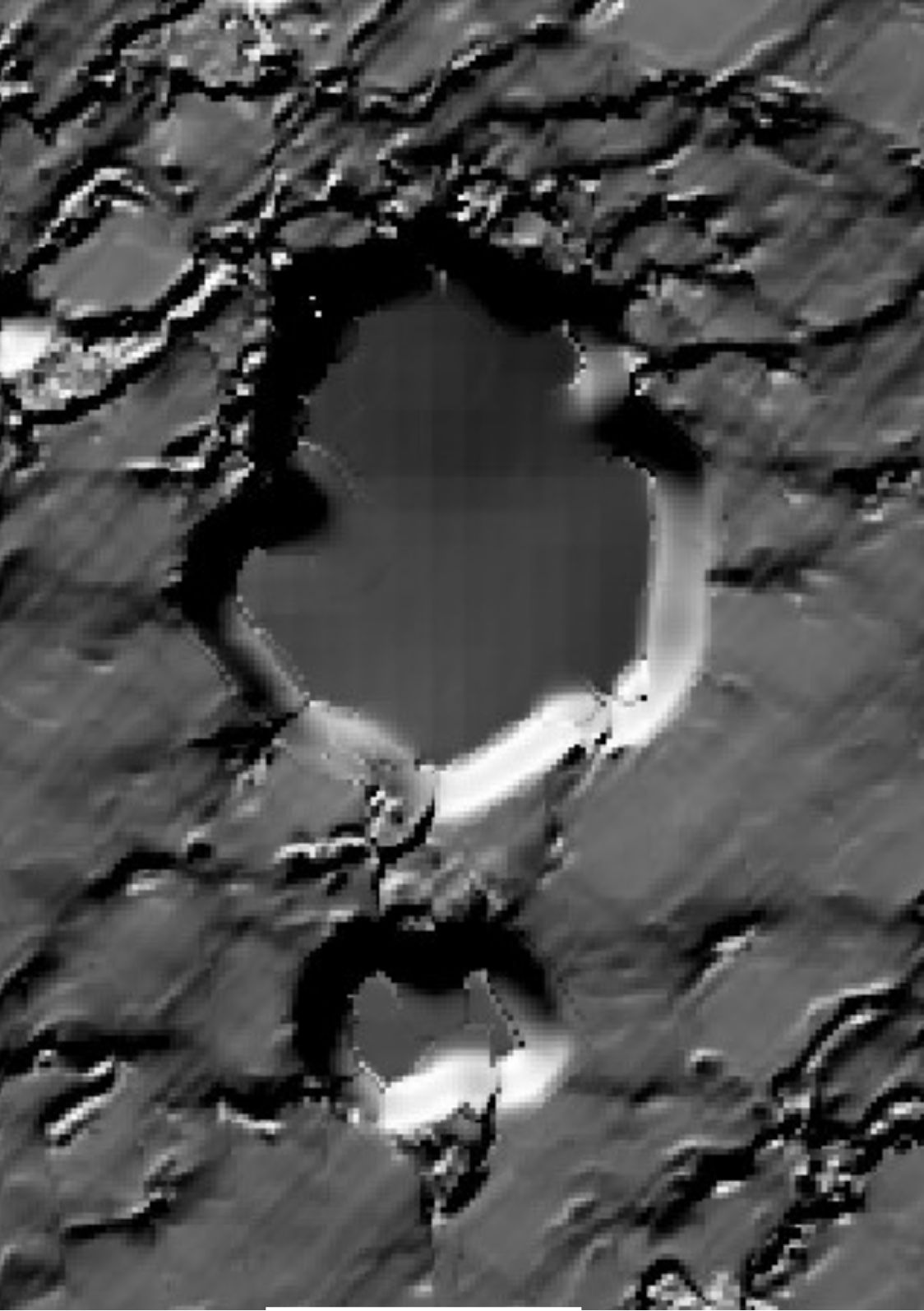}
    \caption{Initial DEM ($z_0$)}
    \label{fig:lc_initial}
  \end{subfigure}
  \hfill
  \begin{subfigure}[b]{0.48\columnwidth}
     \includegraphics[width=\linewidth, height=4cm, keepaspectratio]{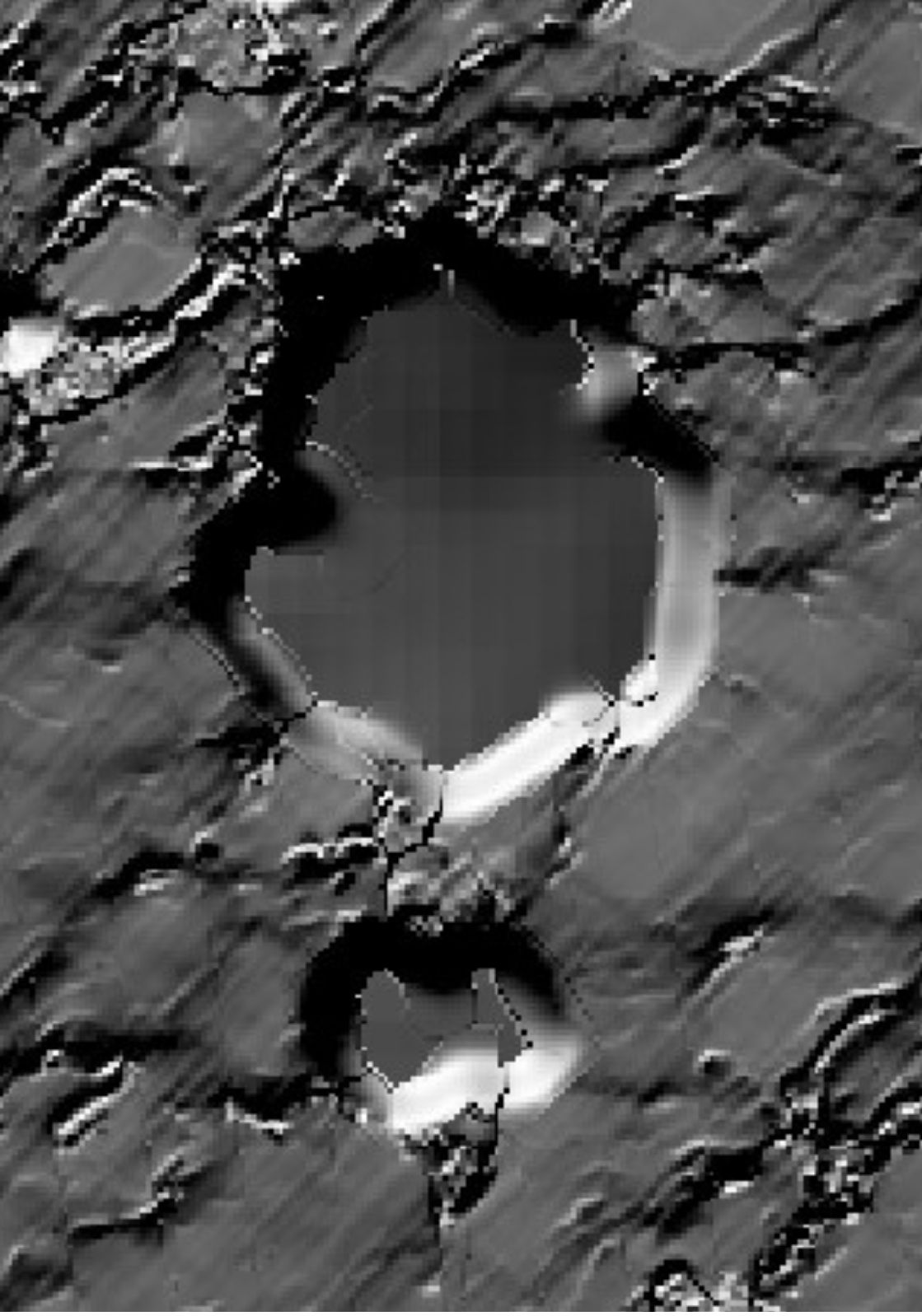}
    \caption{SfS DEM ($z_{\mathrm{SfS}}$)}
    \label{fig:lc_sfs}
  \end{subfigure}
  \caption{Limiting case (Dataset~C): hillshaded initial DEM (left) and
    SfS-enhanced DEM (right). Only subtle textural differences are
    visible between the two panels.}
  \label{fig:limiting_case}
\end{figure}

\textbf{Limiting case: Insufficient illumination diversity and geometric inconsistency.}
Figure~\ref{fig:limiting_case} shows Dataset~C (a small region within the Cyrillus crater), where the small angular separation $\Delta\theta_i$ between $\mathcal{I}_3$ and the stereo pair leads to minimal illumination variation. Only subtle hillshade differences between $z_0$ and $z_{\mathrm{SfS}}$ are observed, with $\Delta\sigma_s$ below 3\% for all $W$ values. This is consistent with the sensitivity of SfS to $\partial R_{\mathrm{LL}} / \partial \theta_i$, where similar illumination conditions limit shading gradients and height updates.

However, Dataset~A, with similar $\Delta\theta_i$, yields stronger reconstruction improvements. This is due to the more consistent viewing geometry in Dataset~A, with a moderate $\Delta\theta_p$, allowing SfS refinement to build on a reliable stereo initialization. In contrast, Dataset~C has a large $\Delta\theta_p \approx 30^\circ$, introducing geometric inconsistencies like foreshortening and shadow projection mismatches, reducing SfS effectiveness.

These results suggest that SfS performance is governed by both illumination diversity and viewing geometry. A combined criterion based on $\Delta\theta_i$ and $\Delta\theta_p$ is more suitable for selecting $\mathcal{I}_3$ than using incidence angle alone.

\textbf{Parameter recommendations.}
For OHRC imagery with adequate $\mathcal{I}_3$ angular separation and
footprint coverage, this study recommends $W = 50$, $C = 10^{-3}$, $N \geq 8$
as the operational parameter set. The decade-then-refine sweep strategy
requires of order 20--30 individual SfS runs and is feasible in a few
CPU-hours per crop using ASP's \texttt{sfs}; \texttt{parallel\_sfs} is
recommended for crops exceeding $\sim$1000$\times$1000 pixels.

\textbf{Pipeline foundation.}
The OHRC stereo reconstruction pipeline follows our prior work~\cite{arxiv_ohrc},
which addressed data-specific challenges in ISIS ingestion and SPICE-based
sensor initialization.

\textbf{Limitations and Future Work.}
The adopted Lunar-Lambertian reflectance model~(\ref{eq:lunar_lambert})
does not capture higher-order photometric effects such as opposition
surge and macroscopic surface roughness, which may become significant
at high incidence angles. In addition, spatial variations in surface
albedo can be misinterpreted as topographic signal in the SfS inversion,
introducing ambiguity in the recovered surface. The current formulation
also relies on a single shading image, making it sensitive to partial
footprint coverage and limiting the robustness of the reconstruction.

A natural extension of this work is the incorporation of multi-image
SfS formulations, which can better constrain the solution under varying
illumination conditions. Furthermore, explicit fusion of the
stereo-derived DEM $z_0$ with the SfS-refined surface offers a promising
direction for improving both absolute accuracy and fine-scale detail.
Such hybrid stereo--photoclinometry approaches have been demonstrated
for LROC NAC datasets within the ASP framework, but are not yet fully
supported for Chandrayaan-2 OHRC processing. Extending the current
pipeline to enable this coupled stereo--SfS integration remains an
important direction for future work.

\section{Conclusion}

This study presents an open-source SfS framework for sub-metre lunar
topographic mapping using Chandrayaan-2 OHRC imagery across multiple
sites. A complete, scientifically characterized OHRC photogrammetric
pipeline has been established. SfS applied to
a third, independent OHRC image functions as a genuine new-data source,
recovering topographic detail spectrally absent from both the stereo
DEM and coarser reference products, as validated by profile comparisons
against the NAC DTM at the Vikram landing region. At Mons Mouton on
the lunar south pole, the technique
reveals fine-scale flank morphology consistent with the known
flat-topped geometry of the massif. This study has additionally characterized
two limiting conditions, that are partial $\mathcal{I}_3$ footprint coverage
and insufficient angular separation between $\mathcal{I}_3$ and the
stereo pair, that bound the operational envelope of the method.
Together, these results provide a reproducible and scientifically
grounded methodology for exploiting the unique sub-metre imaging
capability of OHRC.

\section*{Acknowledgment}

The authors thank the Director, Space Applications Centre (SAC), ISRO, for
providing the opportunity to undertake this work. The authors are grateful to
the Deputy Director, SIPA, and the Head, PSPD for their encouragement and support.

The authors also thank ISRO for providing open access to the Chandrayaan-2 OHRC
imagery and SPICE kernel archive through the ISSDC PRADAN portal. The authors
also acknowledge the NASA Ames Stereo Pipeline (ASP) team and the USGS ISIS
community for their open-source software and providing the
development support for Chandrayaan-2 data, which were integral to this work.

\bibliographystyle{ieeetr}
\bibliography{references}

\end{document}